\documentclass{article}

\usepackage[final]{corl_2019} %
\usepackage{amsmath,amsfonts,amsthm,bm}
\usepackage{graphicx, subcaption}
\usepackage{listings}
\usepackage{courier}
\usepackage{wrapfig}
\usepackage{enumitem}
\usepackage{todonotes}
\usepackage{hyperref}
\usepackage{physics}
\usepackage{gensymb}
\usepackage{siunitx}
\usepackage{enumitem}
\usepackage{xspace}
\usepackage{wrapfig}
\usepackage{bbold}
\usepackage{float}

\newlength{\bibitemsep}
\newlength{\bibparskip}
\let\oldthebibliography\thebibliography
\renewcommand\thebibliography[1]{%
  \oldthebibliography{#1}%
  \setlength{\parskip}{\bibitemsep}%
  \setlength{\itemsep}{\bibparskip}%
}
\usepackage[compact]{titlesec} 
\titlespacing{\section}{0pt}{*0}{*0}
\def\dsuitetxt{ROBEL\xspace}
\def\dclawtxt{D'Claw\xspace}
\def\dkittytxt{D'Kitty\xspace}

\def\dsuite{\emph{\dsuitetxt}\xspace}
\def\dclaw{\emph{\dclawtxt}\xspace}
\def\dkitty{\emph{\dkittytxt}\xspace}

\def\dsuiteWeb{www.roboticsbenchmarks.org}

\title{ \dsuitetxt: \underline{Ro}botics \underline{Be}nchmarks for \underline{L}earning \\ with Low-Cost Robots}

\author{
  Michael Ahn$^{\dagger}$ \And Henry Zhu$^{\delta}$ \And Kristian Hartikainen$^{\delta}$ \And Hugo Ponte$^{\dagger}$  \AND
  Abhishek Gupta$^{\delta}$ \And Sergey Levine$^{\delta\dagger}$ \And  Vikash Kumar$^{\dagger}$ \AND
  $^{\delta}$UC Berkeley, USA \hspace{1cm} $^{\dagger}$Google Research, USA\\
}

\begin{document}
\maketitle

\vspace{-10mm}
\begin{figure}[h!]
  \centering
  \includegraphics[trim=0 0 0 2.35cm, clip, width=1\linewidth]{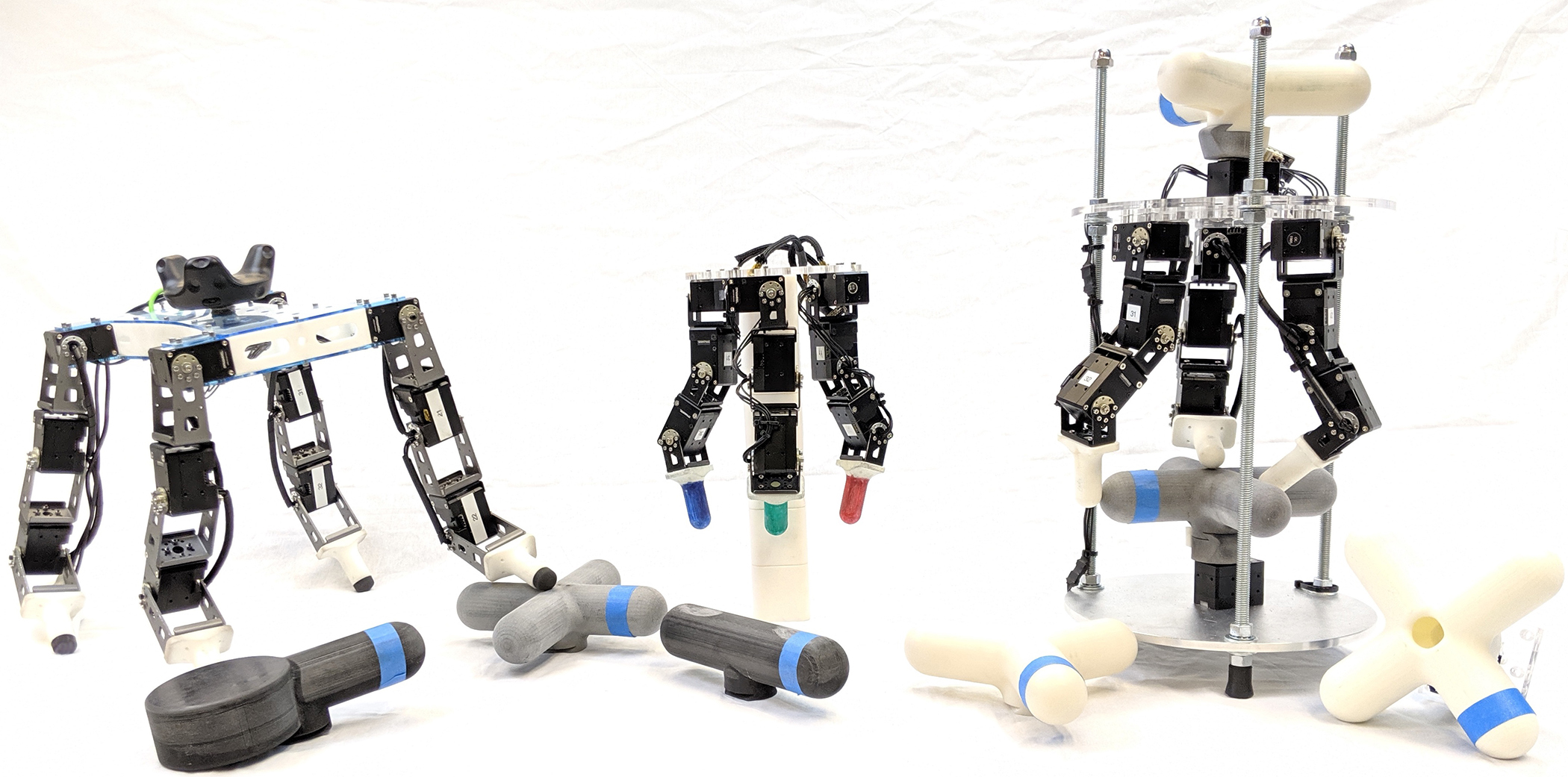}
  \label{fig:cover}
  \vspace{-4mm}
  \caption{\dsuite robots: \dkitty (left) and \dclaw (middle and right)}
\end{figure}
\begin{abstract}
\dsuitetxt is an open-source platform of cost-effective robots designed for reinforcement learning in the real world. \dsuitetxt introduces two robots, each aimed to accelerate reinforcement learning research in different task domains: \dclawtxt is a three-fingered hand robot that facilitates learning dexterous manipulation tasks, and \dkittytxt is a four-legged robot that facilitates learning agile legged locomotion tasks. These low-cost, modular robots are easy to maintain and are robust enough to sustain on-hardware reinforcement learning from scratch with over 14000 training hours registered on them to date. To leverage this platform, we propose an extensible set of continuous control benchmark tasks for each robot. These tasks feature dense and sparse task objectives, and additionally introduce score metrics for hardware-safety. We provide benchmark scores on an initial set of tasks using a variety of learning-based methods. Furthermore, we show that these results can be replicated across copies of the robots located in different institutions. Code, documentation, design files, detailed assembly instructions, trained policies, baseline details, task videos, and all supplementary materials required to reproduce the results are available at \href{\dsuiteWeb}{\dsuiteWeb}

\end{abstract}

\vspace{-4mm}
 
\section{Introduction}

Learning-based methods for solving robotic control problems have recently seen significant momentum, driven by the widening availability of simulated benchmarks \cite{tassa2018deepmind, openAIgym, gym-robotics} and advancements in flexible and scalable reinforcement learning \cite{sutton2018reinforcement, lillicrap2015continuous, peng2017deeploco, haarnoja2018soft}. While learning through simulation is relatively inexpensive and scalable, developments on these simulated environments often encounter difficulty in deploying to real-world robots due to factors such as inaccurate modeling of physical phenomena and domain shift. This motivates the need to develop robotic control solutions directly in the real world on physical hardware.

Modern advancements in reinforcement learning have shown some success in the real world \cite{levine2018learning, pinto2016supersizing, zhu2018dexterous}. However, learning on real robots generally does not take into account physical limitations -- aggressive exploration can induce wear and permanent damage to the robot due to collisions with itself and the surrounding physical environment. A significant portion of current robotics research is conducted on high-cost, industrial-quality robots that are intended for precise, human-monitored operation in controlled environments. Furthermore, these robots are designed around traditional control methods that focus on precision, repeatability, and ease of characterization. This stands in sharp contrast with learning-based methods that are robust to imperfect sensing and actuation, but demand (a) a high degree of resilience to allow real-world trial-and-error learning over a long duration, (b) low cost and ease of maintenance to enable scalability through replication, and (c) reliable mechanisms to allow large scale data collection without strict human monitoring requirements for providing rewards and episodic resets.

To address these emerging requirements, we introduce \dsuite \xspace-- an open-source platform for cost-effective, modular robots that are designed around the needs of reinforcement learning in the real world. This release of \dsuite consists of two robots that are each intended to accelerate research in a distinct task domain: a nine degree of freedom (DOF) manipulation robot \textbf{\dclaw} and a twelve DOF locomotion robot \textbf{\dkitty}. In addition, \dsuite includes a wide variety of benchmark tasks that run in the real world and support a simulated back-end to facilitate rapid prototyping. We present performance metrics on these benchmark tasks over a diverse collection of learning-based methods. Finally, we show that these robots are
replicable and are able to reproduce desired behavior from a control policy that was trained on a different copy of the robot.

\section{Related Work}

Before delving into the specifics of \dsuite, we first review current work looking into simulated benchmarks, the disconnect between the challenges in simulation and reality, hardware benchmarks, and factors influencing real world progress in relevance to continuous control problems in robotics.

Recent advancements in continuous control problems in robotics via learning based methods are fueled in part by access to fast compute at affordable rates, and in part by algorithmic developments \cite{sutton2018reinforcement, allgower2012nonlinear} that generalize and scale well with the complexities of high dimensional problems. Access to easy to use simulated benchmarks \cite{duan2016benchmarking, gym-robotics, tassa2018deepmind} has significantly catalyzed these developments by facilitating fast prototyping, and by providing standard metrics for analysis and comparisons between various methods. 

Various algorithms have been shown to be effective on a large set of simulated environments \cite{peng2017deeploco, haarnoja2018soft}, but these developments have not precipitated down in equal proportions to real world systems, owing to the large divide between the challenges presented by the real world and their simulated counterparts. Precise and programmable resets, noise-free instantaneous observations, high data bandwidth, and lack of concern for environmental safety are a few of the many privileges that the prevalent simulated environments  \cite{duan2016benchmarking, gym-robotics, tassa2018deepmind} enjoy, which are impractical in the real world. In contrast, \dsuite exposes many of these challenges on physical hardware and provides the tools to study them, encouraging future development to directly address these issues.

Algorithms for applying RL to real world robotics have either (a) resorted to solving problems in simulation and transferring to the real world, relying on algorithms like domain randomization to deal with domain shift by solving a more challenging robust control problem~\cite{tobin2017domain,sadeghi2016cad2rl,andrychowicz2018learning,matas2018sim}, or (b) have resorted to completely pose and solve the problem in the real world \cite{pinto2016supersizing, qtOpt}. While the former does not scale well as the task complexity grows \cite{ramos2019bayessim, mehta2019active}, the latter has traditionally required a significant time and cost investment that is task-specific, and is not commonly accessible more broadly to the field. Shared investments in terms of competitive challenges \cite{johnson2015team, behnke2006robot, thrun2006stanley} have also been investigated to boost research and developments on physical robots in the real world. These challenges have failed to stay relevant to the scientific community owing to significant costs and reproducibility issues. \dsuite alleviates these investments by providing a low-cost, easily extensible platform to facilitate real world results by the broader community under reproducible settings.

Roboticists have long been fascinated by the idea of building low cost manipulation \cite{dollarhand, softHand, xu2016design} as well as locomotion platforms \cite{passivewalker, minicheetah}. Many of these platforms can be limited to few DoFs \cite{dollarhand}, aggressively under-actuated \cite{softHand} for simplicity and cost gains, or are difficult to independently assemble and replicate \cite{xu2016design, softHand}. \dsuite leverages its modular design to provide high DOF, easy-to-assemble robots, while retaining easy control and reasonable precision: \dclaw has nine actuated DOFs while remaining low cost. On the other end, \cite{minicheetah} is perhaps the closest, but more expensive, counterpart of our \dkitty platform.

Although not posed as benchmarks, the idea of comparing progress in the real world via shared datasets \cite{ycb}, testbeds \cite{pickem2017robotarium}, and hardware designs \cite{softHand, PR2,lum2009raven} has been around for a while. Recently, benchmarking in the real world using commercially available platforms has also been proposed \cite{mahmood2018benchmarking, yang2019replab}. These benchmarks include robot-centric tasks such end-effector reaching, joint angle tracking, and grasping via parallel jaws grippers. To further diversify the benchmarking scene, \dsuite presents a wide variety of high DoF tasks spanning dexterous manipulation as well as quadruped locomotion.

With learning-based methods \cite{sutton2018reinforcement}, it is common to measure the average episodic return to evaluate the performance of an agent. These returns are task-specific and often ignore the challenges of the real world, such as unsafe exploration, movement quality, hardware risks, energy expenditure, etc. These challenges are highlighted by the DARPA Robotics Challenge \cite{johnson2015team, behnke2006robot, thrun2006stanley} where many robots failed to achieve their task objective due to undervaluing safety objectives, thereby indicating that real world challenges (such as safety) are important objectives to prioritize.  Hardware safety considerations have been posing as explicit constraints (position, velocity, acceleration, jerk limits) as well as regularization (energy, control cost) before, but have not found appropriate emphasis in existing \cite{gym-robotics, tassa2018deepmind, duan2016benchmarking} learning related benchmarks. Addressing this, \dsuite provides three signals (dense-reward, sparse-score and hardware-safety) to facilitate the study of these challenges.

\section{\dsuitetxt} %

\subsection*{Hardware Platforms} %
\begin{wrapfigure}{r}{0.4\textwidth}
    \vspace{-28pt} 
    \centering
    \includegraphics[width=.8\linewidth]{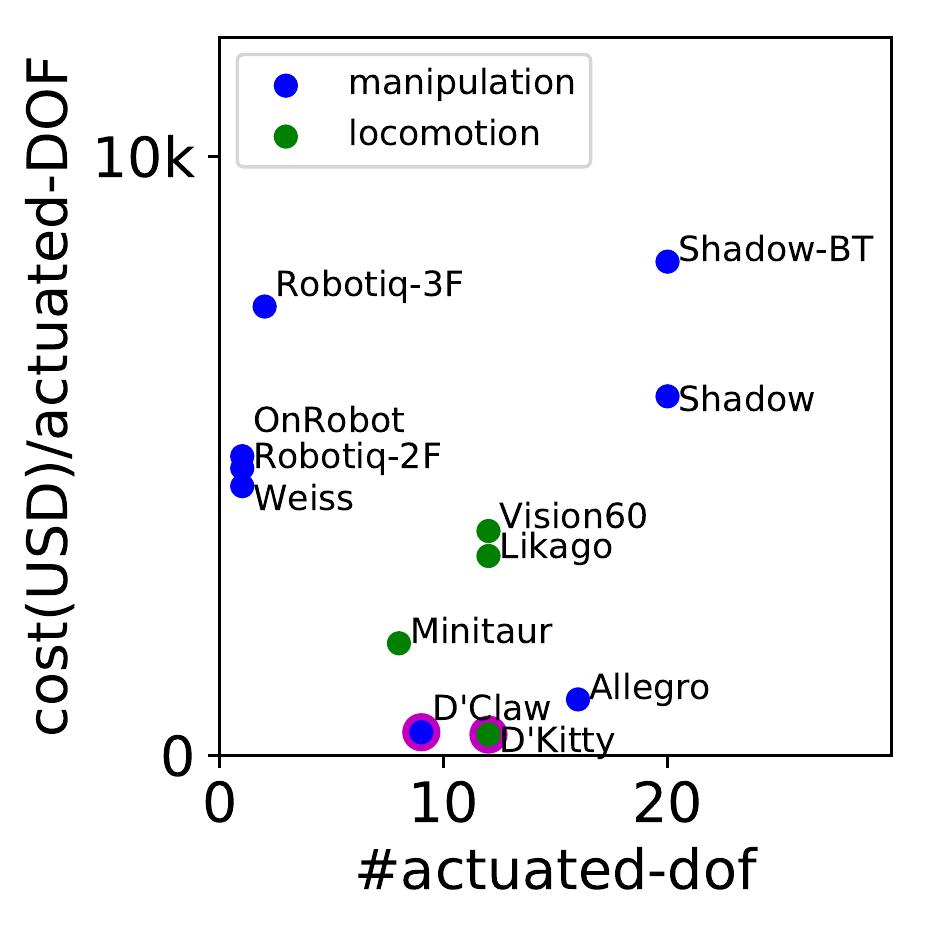}
    \label{fig:cost-landscape}
    \caption{Cost comparison of \dsuite with other commonly used platforms. We note that (a) \dsuite platforms have the most economical price point, thereby facilitating experiment's scalability and (b) Prices scale linearly with \# of DOFs, thanks to modular design, thereby facilitating experiments' complexity}
\end{wrapfigure}

As the number of actuated DOFs of a system grows, we tend to see a proportional increase in cost and decrease in reliability. The modularity of \dsuite allows us to build reasonably high DOF robots while remaining low-cost and easily maintainable. The robots only use off-the-shelf components, commonly-available prototyping tools (3D printers, laser cutters), and require only a few hours to build (\autoref{Tab:prices-build}). \dsuite robots are actuated at joint level (i.e. no transmission between joint and actuators) via \textit{Dynamixel} smart actuators \cite{dynamixel} that feature fully integrated motors with an embedded controller, reduction drive, and high-baudrate communication. Multiple actuators can be daisy-chained together to increase the number of DOFs in the system, which allows \dsuite robots to be easy to build (\autoref{Tab:prices-build}) and extend. For the context of this work we use a USB-serial bus \cite{u2d2} for communication to the robots. An 12V power supply is used to power the platforms. \dsuite platforms also support a wide variety of choices in sensing and actuation modes, which are summarized in \autoref{Tab:capabilities}. 

    \begin{table}[!htb]
        \begin{minipage}{.34\linewidth}
            \begin{flushleft}
            \caption{\dsuite platform initial cost and time investments}
            \label{Tab:prices-build}
            \label{fig:platform-dof-price}
            \begin{tabular}{|l|l|l|}
             \hline
            \textbf{platform} & \textbf{D'Claw} & \textbf{D'Kitty} \\ \hline
                \# DOF            & 9                & 12              \\
                Price (\$)        & 3500             & 4200            \\
                Build(hr)        & 4                & 6               \\ \hline
            \end{tabular}
            \end{flushleft}
        \end{minipage}\hspace{15pt}
        \begin{minipage}{.62\linewidth}
          \begin{flushright}
            \caption{\dsuite platform features a variety of sensing options, control modes, limits and communication speeds}
            \label{Tab:capabilities}
            \begin{tabular}{|l|l|}
            \hline
            \textbf{Property} & \textbf{Options}                                \\ \hline
            Control           & Torque, Velocity, Position, Extended   \\
                              & Position, Current, PWM                          \\
            Sensing           & Position, Velocity, Current, Realtime tick,     \\
                              & Trajectory, Temperature, Input Voltage          \\
            Limits            & Position, Velocity, PWM, Current        \\
            Bandwidth         & 9600 bps $\sim$ 4.5 Mbps                        \\ \hline
            \end{tabular}
            \end{flushright}
        \end{minipage} 
    \end{table}

The schematic details of \dsuite platforms are summarized in \autoref{fig:platforms}. Detailed CAD models and bill of materials (BOM) with step-by-step assembly instructions are included in the supplementary materials package. \dsuite platforms have also been independently replicated and tested for reliability (\autoref{sec:Reliability}) at a geographically remote location which demonstrates the reproducibility (details in \autoref{sec:reproducibility}) of the \dsuite platforms and associated results.

The combination of reproducibility and scalibility exhibited by \dsuite platforms presents to the field of robotics a lucrative preposition of a standard set of benchmarks (proposed in \autoref{sec:Tasks}) to facilitate sharing and collaborative comparison of results. \dsuite consists of the following two platforms:
 
        \begin{figure}[h]
          \centering
          
          \begin{subfigure}[t]{.2\linewidth}
            \centering\includegraphics[width=.9\linewidth]{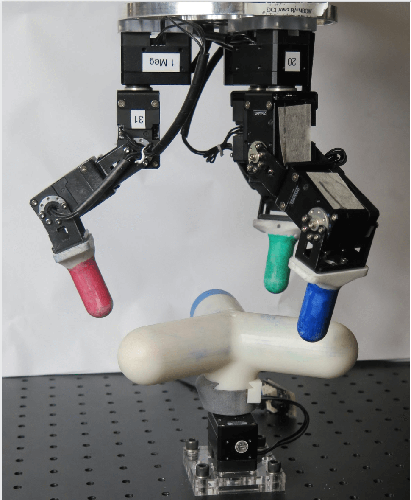}
            \caption{\dclaw-real}
            \label{fig:dclaw-real}
          \end{subfigure}
          \begin{subfigure}[t]{.2\linewidth}
            \centering\includegraphics[width=.9\linewidth]{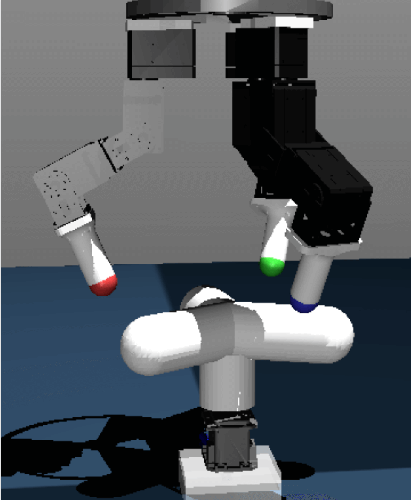}
            \caption{\dclaw-sim}
            \label{fig:dclaw-sim}
          \end{subfigure}
          \begin{subfigure}[t]{.55\linewidth}
            \centering\includegraphics[width=1\linewidth]{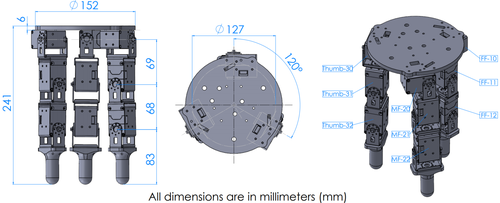}
            \caption{\dclaw schematic details}
            \label{fig:dclaw-sch}
          \end{subfigure}
          \vspace{5pt}
          
          \begin{subfigure}[t]{.2\linewidth}
            \centering\includegraphics[width=1\linewidth]{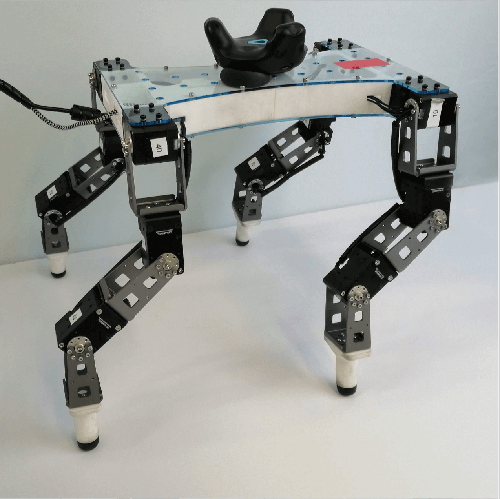}
            \caption{\dkitty-real}
            \label{fig:dkitty-real}
          \end{subfigure}
          \begin{subfigure}[t]{.2\linewidth}
            \centering\includegraphics[width=1\linewidth]{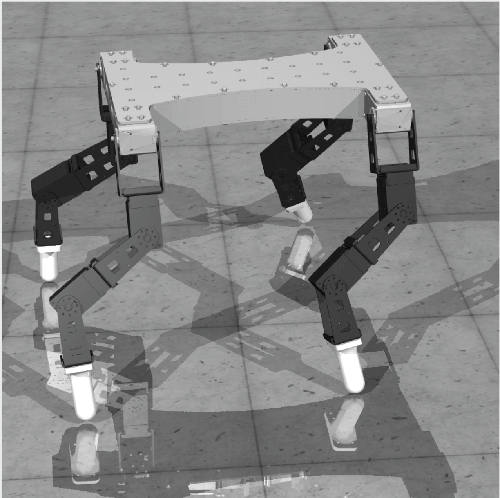}
            \caption{\dkitty-sim}
            \label{fig:dkitty-sim}
          \end{subfigure}
          \begin{subfigure}[t]{.55\linewidth}
            \centering\includegraphics[width=.9\linewidth]{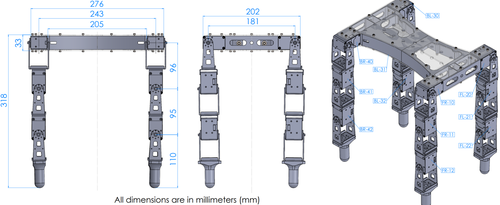}
            \caption{\dkitty schematic details}
            \label{fig:dkitty-sch}
          \end{subfigure}
        
          \caption{\dsuite features two low-cost, robust, modular platforms -- \dclaw (9 DOF manipulation platform) and \dkitty (12 DOF locomotion platform)}
          \label{fig:platforms}
    
        \end{figure}
 \begin{enumerate}
     \item \textbf{\dclaw} (\autoref{fig:dclaw-real}-\ref{fig:dclaw-sch}) is a nine-DOF manipulation platform capable of performing dexterous manipulation. It consists of three identical fingers mounted symmetrically on a circular laser cut base. The finger tips are 3D printed parts. The base can be fixed to a stationary position, or mounted to a portable frame. \dclaw robots have been featured in wide stream of prior research \cite{zhu2018dexterous, haarnoja2018soft, lee2019slac} and have registered 1000s of hours of real world training on them. 
     
     \item \textbf{\dkitty} (\autoref{fig:dkitty-real}-\ref{fig:dkitty-sch}) is a twelve-DOF quadruped capable of agile locomotion. It consists of four identical legs mounted symmetrically on a square base. The feet are simple 3d printed parts with rubber ends. \dkitty is symmetric along all three axes and can also walk normally when upside down.
 \end{enumerate}

\section{Benchmark Tasks}
\label{sec:Tasks}

\dsuite proposes a collection of tasks for \dclaw and \dkitty to serve as a foundation for real-world benchmarking for continuous control problems in robotics. We first outline the formulations of these benchmark tasks, and then provide details of the tasks grouped into manipulation and locomotion.

\dsuite tasks are formulated in a standard Markov decision process (MDP) setting~\cite{puterman1994markov}, in which each step, corresponding to a time $t$ in the environment, consists of a state \textit{observation} $s$, an input \textit{action} $a$, a resulting \textit{reward} $r_d$, and a resulting \textit{next state} $s'$. In addition to the reward $r_d$, which is usually dense, \dsuite also provides a sparse signal called \textit{score} $r_s$, which can be interpreted as a sparse task objective without any shaping. To standardize quantification a \textit{policy}'s $\pi$ performance, \dsuite provides \textit{success evaluator} $\phi_{se}(\pi)$ metrics and \textit{hardware safety} $\phi_{hs}(\pi)$ metrics.

\begin{wrapfigure}{r}{0.4\textwidth}
\vspace{-10pt}
\centering
\includegraphics[width=\linewidth]{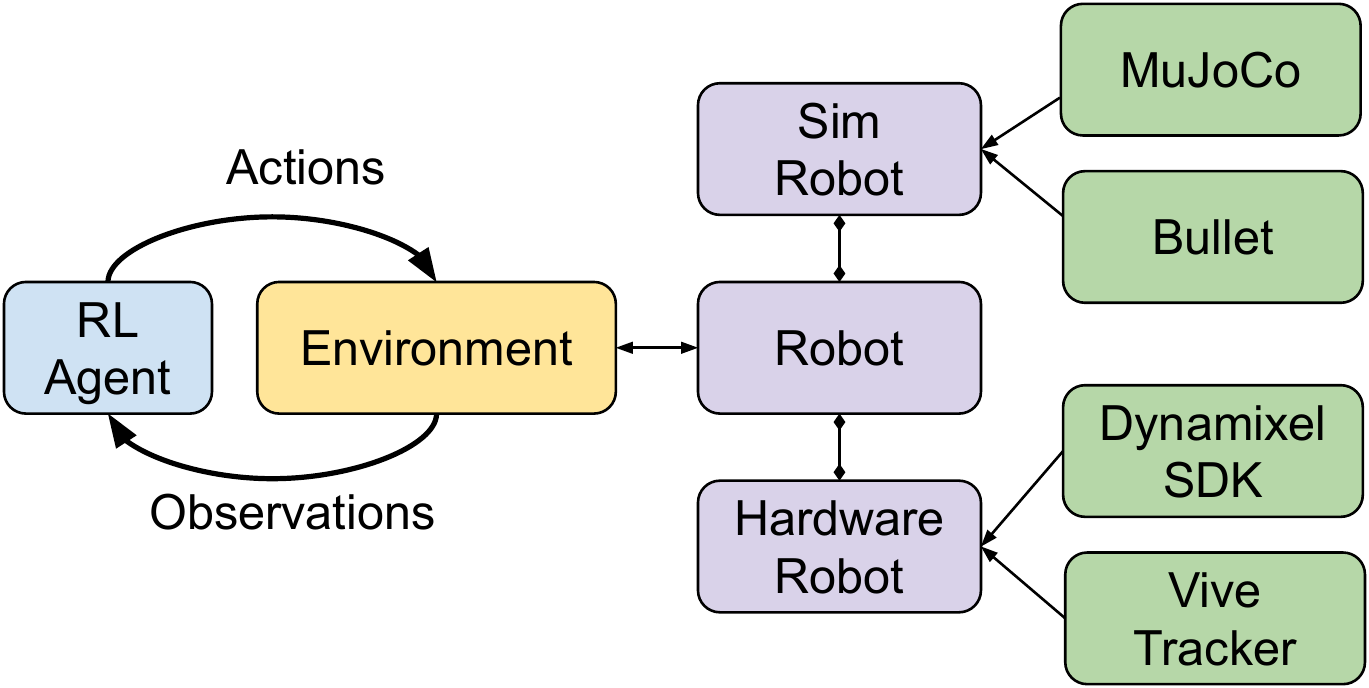} 
\caption{\dsuite architecture}
\label{fig:Architecture}
\end{wrapfigure}

To implement the MDP setting, we employ the commonly-adopted OpenAI Gym \cite{openAIgym} API. \dsuite is presented as an open-source Python library consisting of modular, reusable software components that enable a common interface to interact with hardware and simulation. \autoref{fig:Architecture} provides architectural outline of \dsuite. The implementation of the environments are largely agnostic to whether they are running on real \textit{hardware robot} or \textit{simulation robot}. The simulated robot is a modular component of the system and can be exchanged for any physics simulation engines. \autoref{fig:dclaw-sim}, and \autoref{fig:dkitty-sim} show the simulated robot modelled in MuJoCo \cite{todorov2012mujoco}. We encourage the usage of simulation primarily as a rapid prototyping tool and promote purely real-world hardware results as \dsuite benchmarks.

The \textit{reward} $r_d$ is the most commonly used signal in reinforcement learning that the agents directly optimize. Since the reward often consists of multiple sub-goals and regularization terms, \textit{score} $r_s$ provides a more direct task-specific sparse objective. \textit{Success evaluator} $\phi_{se}(\pi)$ is defined to be reward (or score) agnostic. It evaluates success (task-specific) percentage of policy over multiple runs. Unlike rewards and score, which are provided at each step, \textit{hardware safety} $\phi_{hs}(\pi)$ is an array of counters that evaluates a policy over the specified horizon to measure the number of safety violations. We include the following violations in our safety measure: joint limits, velocity limits, and current limits.

We propose an initial set of \dsuite benchmark tasks to tackle a variety of challenges involving manipulation and locomotion. We summaries the tasks below and encourage readers to refer to \autoref{app:task-details} and supplementary material\footnote{\label{supplimentary} code repository, detailed documentation, and task videos are available at  \href{\dsuiteWeb}{\dsuiteWeb}} for task details. 

\subsection{Manipulation Benchmarks on D'Claw}
\label{sec:dclaw-tasks}
\dclaw is 9-DoF dexterous manipulator capable of contact rich diverse behaviors. We structure our first group of the benchmark tasks around fundamental manipulation behaviors.

\begin{figure}[h]
    \centering
    
    \begin{subfigure}[t]{.32\linewidth}
        \centering\includegraphics[width=1\linewidth]{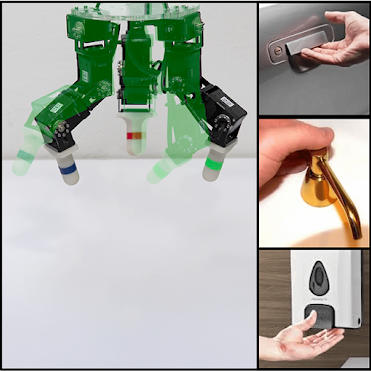}
        \caption{Pose: conform to a shape}
        \label{fig:dclaw-task-pose}
    \end{subfigure}
    \begin{subfigure}[t]{.32\linewidth}
        \centering\includegraphics[width=1\linewidth]{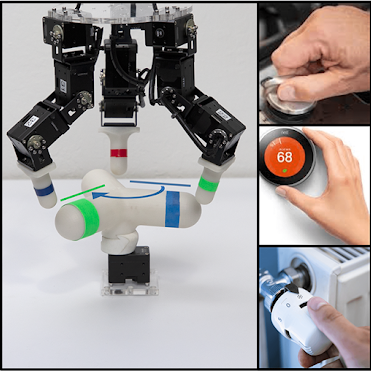}
        \caption{Turn: rotate to a fixed target}
        \label{fig:dclaw-task-turn}
    \end{subfigure}
    \begin{subfigure}[t]{.32\linewidth}
        \centering\includegraphics[width=1\linewidth]{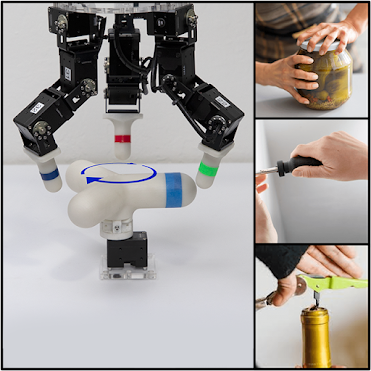}
        \caption{Screw: rotate to a moving target}
        \label{fig:dclaw-task-screw}
    \end{subfigure}
    \caption{\dclaw manipulation benchmarks: Pose, Turn and Screw are motivated by commonly observed manipulation behaviors in daily life}
    \label{fig:dclaw-challenges}
\end{figure}

\renewcommand{\labelenumi}{\alph{enumi})}
\begin{enumerate}[wide, labelwidth=*, labelindent=0pt]
    \item \textbf{Pose} (conform to the shape of the environment): This task is motivated by the primary objective of a manipulator to conform to its surrounding in order to prepare for the upcoming maneuvers -- commonly observed as various pre-grasp and latching maneuvers (\autoref{fig:dclaw-task-pose}). This set of tasks is posed as trying to match randomly selected joint angle targets. Successful completion of this task demonstrates the capability of a manipulator to have controlled access to all its joints. This set of tasks are comparatively easier to train, thereby facilitates fast iteration cycles and a gradual transition to the rest of the tasks. %
    Two variants of this task are provided: 
    a static variant \textit{DClawPoseFixed} where the desired joint angles remain constant, and a dynamic variant \textit{DClawPoseRandom} where the desired joint angle is time-dependent and oscillates between two goal positions that are sampled at the beginning of the episode. %
    
    \item \textbf{Turn} (rotate to a fixed target angle): This task encapsulates the ability of a manipulator to reposition unactuated DoFs present in the environments to target configurations -- commonly observed as turning various knobs, latches and handles. This set of tasks is posed as trying to match randomly selected joint angle targets for the unactuated object(s). Successful completion of this task demonstrates the ability of a manipulator to bring desired changes on external targets. In order to succeed, the manipulator requires not only co-ordination between the internal DoFs, but also an understanding of environment dynamics perceived through contact interactions. Three variants of this task are provided: 
    \textit{DClawTurnFixed} where initial and target angles are constant,  
    \textit{DClawTurnRandom} where both initial and target angles are randomly selected,
    \textit{DClawTurnRandomDynamics} where initial and target angles are randomly selected as well as the environment (object size, surface, and dynamics properties) is randomized. %
    
    \item \textbf{Screw} (rotate to a moving target angle): This task focuses on the ability of a manipulator to continuously rotate an unactuated object at a constant velocity. This set of tasks is posed as trying to match joint angle targets that are themselves moving. Although very similar to turn tasks but the nuances of moving target challenge the manipulator's strategy to constantly evolve as the target drifts. Fingers often enter singular positions as the rotation progresses. A successful strategy needs to learn finger co-ordinated gating to simultaneous progress as well as stay out of local minima. Three variants of this task are provided: 
    \textit{DClawScrewFixed} where target velocity is constant,  
    \textit{DClawScrewRandom} where the initial angle and target velocity is randomly selected,
    \textit{DClawScrewRandomDynamics} where the initial angle and target velocity is randomly selected as well as the environment (object size, surface, and dynamics properties) is randomized. %
\end{enumerate}

\subsection{Locomotion Benchmark on D'Kitty}
The twelve DoF locomotion platform \dkitty is capable of exhibiting diverse behaviors. We structure this group of the benchmark tasks on the platform around simple locomotion behaviors exhibited by quadrupeds.

  \begin{figure}[h]
  \vspace{-8pt}
    \setlength{\fboxsep}{0pt}%
    \setlength{\fboxrule}{1pt}%
          \centering
          
          \begin{subfigure}[t]{.32\linewidth}
            \centering\fbox{\includegraphics[width=.95\linewidth]{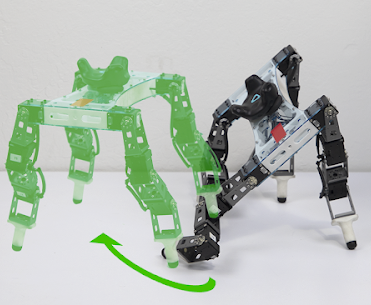}}
            \caption{Stand: getting upright}
            \label{fig:dkitty-stand}
          \end{subfigure}
          \begin{subfigure}[t]{.32\linewidth}
            \centering\fbox{\includegraphics[width=.95\linewidth]{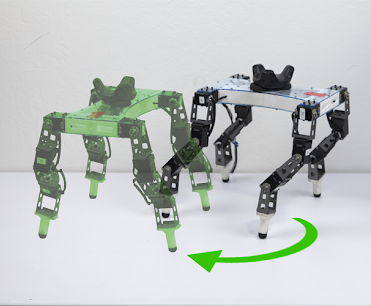}}
            \caption{Orient: align heading}
            \label{fig:dkitty-orient}
          \end{subfigure}
          \begin{subfigure}[t]{.32\linewidth}
            \centering \fbox{\includegraphics[width=.95\linewidth]{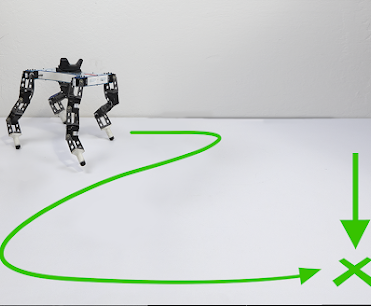}}
            \caption{Walk: get to target}
            \label{fig:dkitty-walk}
          \end{subfigure}
          
          \caption{\dkitty locomotion benchmarks}
          \label{fig:dkitty-challenges}
        
        \end{figure}
        
\renewcommand{\labelenumi}{\alph{enumi})}
\begin{enumerate}[wide, labelwidth=*, labelindent=0pt]
    \item \textbf{Stand}: Standing upright is one of the most fundamental behavior exhibited by the animals. This task involves reaching a pose while being upright. A successful strategy requires maintaining the stability of the torso via the ground reaction forces. Three variants of this task are provided: 
    \textit{DKittyStandFixed} standing up from a fixed initial configuration, 
    \textit{DKittyStandRandom} standing up from a random initial configuration, 
    \textit{DKittyStandRandomDynamics} standing up from random initial configuration where the environment (surface, dynamics properties of \dkitty and ground height map) is randomized. See supplementary materials\textsuperscript{\ref{supplimentary}} for full details
    
    \item \textbf{Orient}: This task involves \dkitty changing its orientation from an initial facing direction to a desired facing direction. This set of tasks is posed as matching the target configuration of the torso. A successful strategy requires maneuvering the torso via the ground reaction forces while maintaining balance. Three variants of this task are provided: 
    \textit{DKittyOrientFixed} maneuvers to a fixed target orientation, 
    \textit{DKittyOrientRandom} maneuvers to a random target orientation, 
    \textit{DKittyOrientRandomDynamics} maneuver to a random target orientation where the environment (surface, dynamics properties of \dkitty and ground height map) is randomized. See supplementary materials\textsuperscript{\ref{supplimentary}} for full details
    
    \item \textbf{Walk}: This task involves the \dkitty moving its world position from an initial cartesian position to desired cartesian position while maintaining a desired facing direction. This task is posed as matching the cartesian position of the torso with a distant target. Successful strategy needs to exhibit locomotion gaits while maintaining heading.  Three variants of this task are provided: 
    \textit{DKittyWalkFixed} walk to a fixed target location, 
    \textit{DKittyWalkRandom} walk to a randomly selected target location, 
    \textit{DKittyWalkRandomDynamics} walk to a selected target location where the environment (surface, dynamics properties of \dkitty and ground height map) is randomized. See supplementary materials\textsuperscript{\ref{supplimentary}} for full details

\end{enumerate}

\dsuite tasks variants are carefully designed to represent a wide task spectrum. The \textit{fixed} variants (task-name suffix ``Fixed'') are fast to iterate and are helpful for getting started. The \textit{random} variants (task-name suffix ``Random'') present a wide initial and goal distribution to study task generalization. In addition to the wider distribution, the \textit{random dynamics} variants (task-name suffix ``RandomDynamics'') presents variability in the various environment properties. This variant is hardest to solve and is well suited for sim2real line of research.
\section{Experiments}

We first summarize on-hardware training runs of various reinforcement learning algorithms that are included as \dsuite baselines. Later  we evaluate \dsuite for its reproducibility with-in the same as well as at a geographically separated location, and reliability over extended usage. We conclude by presenting performance of our baselines over the proposed safety metrics.

\subsection{Baselines}
\vspace{-5pt}

    \begin{figure}[t!]
      \centering
        \includegraphics[width=1\linewidth]{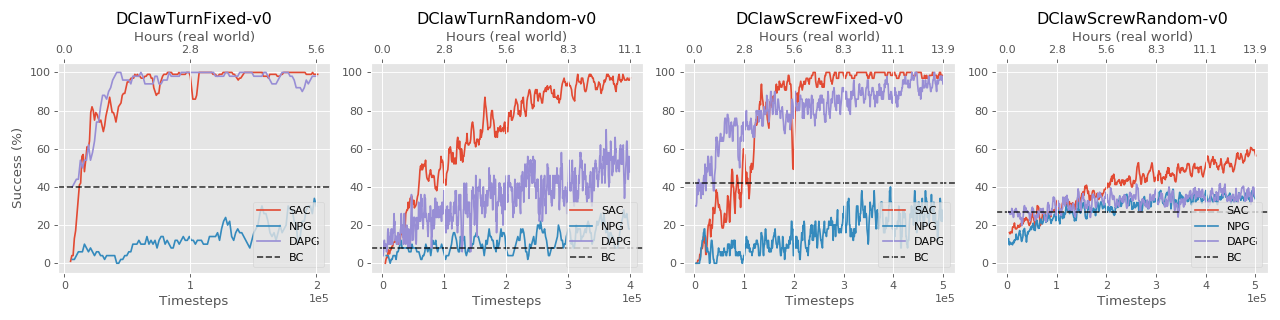}
        \includegraphics[width=1\linewidth]{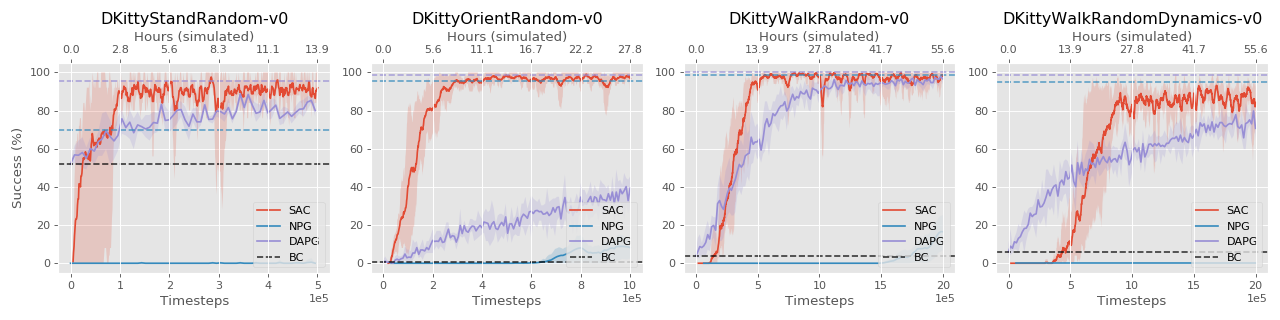}
        \caption{
        Success percentage for \dclaw and \dkitty tasks trained on a physical \dclaw robot and a simulated \dkitty robot using several agents: Soft Actor Critic (SAC)\cite{haarnoja2018soft}, Natural Policy Gradient (NPG)\cite{kakade2002natural}, Demo-augmented Policy Gradient (DAPG)\cite{rajeswaran2017learning}, and Behavior Cloning (BC) over 20 trajectories. Success is measured via the success evaluator $\phi_{se}(\pi)$ of the task (See \autoref{app:task-details} for details). Each timestep corresponds to $0.1$ real-world seconds}
      \label{fig:dsuite-baselines}
      \label{fig:dkitty-baselines}
    \end{figure}

\dsuite has been tested to meet the rigor of a wide variety of learning algorithms. One candidate from each algorithmic class was added to the spectrum of baselines (\autoref{fig:dsuite-baselines}). We include Natural Policy Gradient \cite{kakade2002natural} for on-policy, Soft Actor Critic \cite{haarnoja2018soft} for off-policy, Demo Augmented Policy Gradients \cite{rajeswaran2017learning} for demonstration accelerated methods, and behavior cloning as supervised learning baseline. Using the sim-robot, dynamics randomized variant of all the tasks (referred as randomDynamics) are also included in the package to facilitate sim2real research direction. We also invite the open source community to add to our family of baselines via our open source repository.

\subsection{Reproducibility}
\label{sec:reproducibility}
\vspace{-5pt}

\begin{figure}[h]
  \vspace{-5pt}
  \centering
    \includegraphics[width=1\linewidth]{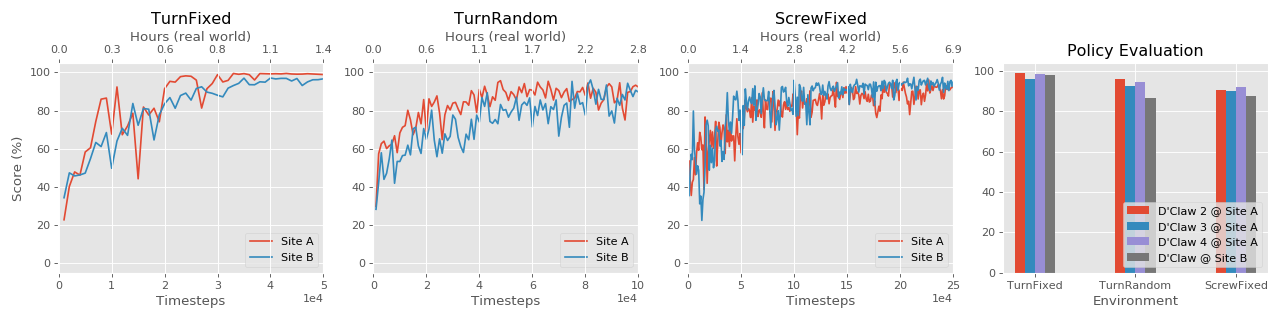}
    \caption{Left-3: Training reproducibility between two real \dclaw robots, developed at different laboratory locations, over the benchmark tasks. Right: Effectiveness of a policy on different hardware it wasn't trained on. Score denotes closeness to to the goal.}
  \label{fig:dsuite-reproduce}
\end{figure}
        
We test \dsuite reproducibility on multiple platforms independently developed at different locations (60 miles apart) via different groups (no in-person visits) using only \dsuite documentation\footnote{Occasional minor clarification over emails were later adopted into the documentation}. We evaluate \dsuite's reproducibility by studying the effectiveness of a policies on different hardware. \autoref{fig:dsuite-reproduce} outlines the effectiveness of a policy on multiple hardware across two different sites. %

\subsection{Reliability}
\label{sec:Reliability}
\vspace{-5pt}
We provide a qualitative measure of the reliability of the system in \autoref{table:reliability}. It should be noted that these metrics include  data gathered while the system was under development. The matured system reported in this paper is much more reliable. \autoref{fig:new-used} provides a qualitative depiction of the robustness of the system using side by side comparison of a new and used \dclaw assembly. The system is fairly robust in facilitating multiple day real-world experimentation on the hardware. Occasional maintenance needs primarily include screws becoming loose. We attribute this to the vibrations caused by recurring collision impacts during manipulation and locomotion. We also observe occasional motor failures (\autoref{table:reliability}). Owing to the modularity of \dsuite, this is easy to replace\footnote{Broken motors are repairable via manufacturers RMA. Motor sub-assemblies are available online as well.}.
\vspace{-2mm}

  \begin{minipage}{\textwidth}
      \begin{minipage}{0.32\textwidth}
        \centering
        \includegraphics[width=.95\linewidth]{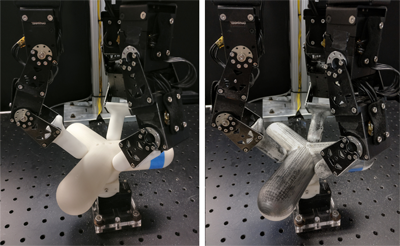}\vspace{6pt}
      \end{minipage}\hfill
      \begin{minipage}{0.32\textwidth}
        \centering
        \includegraphics[width=\linewidth]{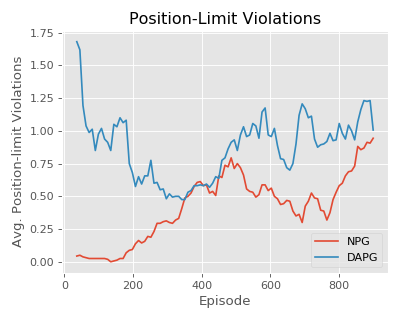}\vspace{2pt}
      \end{minipage}\hfill
      \begin{minipage}{0.32\textwidth}
        \centering
        \begin{tabular}{|l|l|l|}
        \hline
        site                                                     & A    & B    \\ \hline
        \begin{tabular}[c]{@{}l@{}}training\\ hours\end{tabular} & 9000 & 5000 \\ \hline
        \begin{tabular}[c]{@{}l@{}}motors\\ bought\end{tabular}  & 150  & 40   \\ \hline
        \begin{tabular}[c]{@{}l@{}}motors\\ broken\end{tabular}  & 19   & 10   \\ \hline
        \end{tabular}\vspace{4pt}
      \end{minipage}
      
      \begin{minipage}[t]{0.32\textwidth}
          \captionof{figure}{Change in physical appearance depicting \dclaw resilience to extreme usage. (left: new \dclaw) (right: operational for \~6 months)}
          \label{fig:new-used}
      \end{minipage}\hfill
      \begin{minipage}[t]{0.32\textwidth}
        \captionof{figure}{Safety violations observed during the training of the \emph{DClawScrewFixed} task} %
        \label{fig:dclaw-safety}
      \end{minipage}\hfill
      \begin{minipage}[t]{0.32\textwidth}
        \captionof{table}{(approximate) Usage statistics of \dsuite over 12 months. Note that statistics include data from when \dsuite was still under development}
        \label{table:reliability}
      \end{minipage}
  \end{minipage}

\subsection{Safety}
\vspace{-5pt}

Smooth, elegant behavior has been a desirable but hard-to-define trait for all continuous control problems. Various forms of regularization on control, velocity, acceleration, jerks, and energy are often used to induce such properties. While there is not a universally accepted definition for smoothness, few metrics for safe behaviors can be defined in terms of hardware safety limits. In addition to dense and sparse objective, \dsuite also provides hardware safety objectives, which has been largely ignored in available benchmarks \cite{gym-robotics}\cite{duan2016benchmarking}\cite{tassa2018deepmind}. \dsuite defines safety objects over position, velocity, and torque violations calculated over a finite horizon trajectory. The \textit{success evaluator}, provided with all benchmarks, not only reports the average task success metric, but also reports the average number of safety violations. A benchmarks challenge is considered successful when there are no safety violations. \autoref{fig:dclaw-safety} shows the average number of joint per episode under safety violations for two RL agents. We observe that these policies, while successful in solving the task, exhibit significant safety violations. While safety is desirable, it has largely been ignored in existing RL benchmarks resulting in limited progress. We hope that safety-metric included in \dsuite will sprout research in this direction.

\section{Conclusion}

This work proposes \dsuite -- an open source platform of cost-effective robots designed for on-device reinforcement learning experimentation needs. \dsuite platforms are robust and have sustained over 14000 hours of real world training on them till date. \dsuite feature a 9-DOF manipulation platform \dclaw, and a 12-DOF locomotion platform \dkitty~, with a set of prepackaged benchmark tasks around them. We show the performance of these benchmarks on a variety of learning-based agents -- on-policy (NPG), off-policy (SAC), demo-accelerated method (DAPG), and supervised method (BC). We provide these results as baselines for ease of comparison and extensibility. We show reproducibility of the \dsuite's benchmarks by  independently reproducing results at a remote site. We are excited to bring \dsuite to the larger robotics community and look forward to the possibilities it presents towards the evolving experimentation needs of learning-based methods, and robotics in general.

\clearpage
\acknowledgments{
We thank Aravind Rajeswaran, Emo Todorov, Vincent Vanhoucke, Matt Neiss, Chad Richards, Thinh Nguyen, Byron David, Garrett Peake, Krista Reymann, and the rest of Robotics at Google for their contributions and discussions all along the way.
}

\bibliography{references}  %

\newpage
\appendix

\section*{Appendix}

\section{\dsuite task details}
\label{app:task-details}
In this section, we outline details of the benchmark task presented in \autoref{sec:Tasks}.

\subsection{\dclaw tasks}

The action space of all \dclaw tasks (\autoref{sec:dclaw-tasks}) is a 1D vector of 9 \dclaw joint positions.

\renewcommand{\labelenumi}{\alph{enumi})}
\begin{enumerate}[wide, label=(\roman*), labelindent=-5pt, leftmargin=5mm]
    \item \textbf{Pose}:
    This task involves posing \dclaw by driving its joints $\bm{\theta}_t$ to a desired joint angles $\bm{\theta}_{goal}$ sampled randomly from the feasible joint angle space at the beginning of the episode.
    The observation space $s_t $ is a 36-size 1D vector that consists of the current joint angles $\bm{\theta}_t$, the joint velocities $\bm{\dot{\theta}}_t$, the error between the goal and current joint angles, and the last action. The reward function is defined as:
    $$ r_t = -\norm{\bm{\theta}_{goal} - \bm{\theta}_t} - 0.1 \norm{\bm{\dot{\theta}}_t * \mathbb{1}(|\bm{\dot{\theta}}| > 0.5)} $$

    Three variants of this task are provided: 
    \begin{enumerate}
        \item \textit{DClawPoseFixed}: a static variant where the desired joint angles remain constant for the episode
        \item \textit{DClawPoseRandom}: a dynamic variant where the desired joint angle is time-dependent and oscillates between two goal positions that are sampled at the beginning of the episode.
        \item \textit{DClawPoseRandomDynamic}: same as previous. The joint damping, and the joint friction loss are randomized at the beginning of every episode.
    \end{enumerate}
    
    Success evaluator metric $\phi_{se}(\pi)$ of policy $\pi$ is defined using the mean absolute tracking error being within the threshold $\beta = 10\degree$\
     
     $$\phi_{se}(\pi) = \mathbb{E}_{\tau\sim\mathcal{\pi}} \Big[\frac{1}{T} \sum_{t=0}^{T} mean\norm{\theta_{goal}^{(\tau)} - \theta_{t}^{(\tau)}} < \beta \Big]$$

    \item \textbf{Turn}:
    This task involves rotating an object from an initial angle $\theta_{0, obj}$ to a goal angle $\theta_{goal, obj}$. The observation space is a 21-size 1D vector of the current joint angles $\bm{\theta}_{t}$, the joint velocities $\bm{\dot{\theta}}_{t}$, the sine and cosine values of the object's angle $\theta_{t,obj}$, the last action, and the error between the goal and the current object angle $\Delta\theta_{t,obj} = \theta_{t,obj}-\theta_{goal, obj}$. The reward function is defined as
    $$ r_t = -5 |\Delta{\theta_{t,obj}}| - \norm{\bm{\theta}_{nominal} - \bm{\theta}_{t}} - \norm{\bm{\dot{\theta}}_{t}} + 10 \mathbb{1}(|\Delta{\theta_{t,obj}}| < 0.25) + 50 \mathbb{1}(|\Delta{\theta_{t,obj}}| < 0.1) $$
    
    Three variants of this task are provided:
    \begin{enumerate}
        \item \textit{DClawTurnFixed}: constant initial angle ($0\degree$) and constant goal angle ($180\degree$).
        \item \textit{DClawTurnRandom}: random initial and goal angle.
        \item \textit{DClawTurnRandomDynamics}: same as previous. The position of the \dclaw relative to the object, the object's size, the joint damping, and the joint friction loss are randomized at the beginning of every episode.
    \end{enumerate}
    
    Success evaluator metric $\phi_{se}(\pi)$ of policy $\pi$ is defined using the error in last step of the episode $(t=T)$ being within the goal threshold $\beta = 0.1$ as:
    $$ \phi_{se}(\pi) = \mathbb{E}_{\tau\sim\mathcal{\pi}}[\Delta\theta_{T, obj}^{(\tau)} < \beta]$$
    
    \item \textbf{Screw}: This task involves rotating an object at a desired velocity $\dot{\theta}_{desired}$ from an initial angle. This is represented by a $\theta_{t, goal}$ that is updated every step as $\theta_{t, goal} = \theta_{t-1, goal} + \dot{\theta}_{desired}*dt$. Screw tasks have the same observation space and reward definitions as the Turn tasks. Three variants of this task are provided:

    \begin{enumerate}
        \item \textit{DClawScrewFixed}: constant initial angle ($0\degree$) and velocity ($0.5\frac{\text{rad}}{\sec}$)
        \item \textit{DClawScrewRandom}: random initial angle ($[-180\degree, 180\degree]$) and desired velocity ($[-0.75\frac{\text{rad}}{\sec}, 0.75\frac{\text{rad}}{\sec}]$)
        \item \textit{DClawScrewRandomDynamics}: same as previous. The position of the \dclaw relative to the object, the object's size, the joint damping, and the joint friction loss are randomized at the beginning of every episode.
    \end{enumerate}

     Success evaluator metric $\phi_{se}(\pi)$ of policy $\pi$ is defined using the mean absolute tracking error being within the threshold $\beta = 0.1$
     
     $$ \phi_{se}(\pi) = \mathbb{E}_{\tau\sim\mathcal{\pi}}\Big[\frac{1}{T} \sum_{t=0}^T | \Delta\theta_{t,obj}^{(\tau)}|  < \beta \Big]$$
     
\end{enumerate}

\subsection{\dkitty tasks}

The action space of all of the \dkitty tasks is a 1D vector of 12 joint positions. The observation space shares 49 common entries: the Cartesian position (3), Euler orientation (3), velocity (3), and angular velocity (3) of the \dkitty torso, the joint positions $\bm{\theta}$ (12) and velocities $\bm{\dot{\theta}}$ (12) of the 12 joints, the previous action (12), and `uprightness' $u_{t, kitty}$ (1). The uprightness $u_{t, kitty}$ of the \dkitty is measured as it's orientation projected over the global vertical axis:
$$ u_{t, kitty} = \bm{R}_{\hat{z},t, kitty} \cdot \bm{\hat{Z}} $$

The \dkitty tasks share a common term in the reward function $ r_{t, upright}$ regarding uprightness defined as:
$$ r_{t, upright} = \alpha_{upright} \frac{ u_{t, kitty} - \beta }{ 1 - \beta } + \alpha_{falling} ( u_{t, kitty} < \beta )  $$
where $\beta$ is the cosine similarity threshold with the global z-axis beyond which we consider the \dkitty to have fallen. When perfectly upright $\alpha_{t, upright}$ reward is collected, when alignment ($u_{t, kitty}$) falls below the threshold $\beta$, the episode terminates early and $\alpha_{falling}$ is collected.

\renewcommand{\labelenumi}{\alph{enumi})}
\begin{enumerate}[wide, label=(\roman*), labelindent=-5pt, leftmargin=5mm]
    \item \textbf{Stand}: This task involves \dkitty coordinating its 12 joints $\bm{\theta}_t$ to stand upright maintaining a pose specified by $\bm{\theta}_{goal}$. The observation space is a 61-size 1D vector of the shared observation space entries and pose error $\bm{e}_{t, pose} = (\bm{\theta}_{goal} - \bm{\theta}_{t}) $. The reward function is defined as:
    $$ r_t = r_{t, upright} - 4 \bar{e}_{t, pose} - 2 || \bm{p}_{t, kitty} ||_2 + 5 u_{t, kitty} \mathbb{1}(\bar{e}_{t, pose} < \frac{\pi}{6}) + 10 u_{t, kitty} \mathbb{1}(\bar{e}_{t, pose} < \frac{\pi}{12}) $$
    where $\bar{e}_{t, pose}$ is mean absolute pose error, $\bm{p}_{t, kitty}$ is the cartesian position of \dkitty on the horizontal plane and the shared reward function constants are $\alpha_{upright} = 2$, $\alpha_{falling} = -100$, $\beta = \cos(90\degree)$.
    
    Three variants of this task are provided:
    \begin{enumerate}
        \item \textit{DKittyStandFixed}: constant initial pose.
        \item \textit{DKittyStandRandom}: random initial pose.
        \item \textit{DKittyStandRandomDynamics}: same as previous. The joint gains, damping, friction loss, geometry friction coefficients, and masses are randomized. In addition, a randomized height field is generated with heights up to $0.05\si{m}$
    \end{enumerate}
    
    The successor evaluator indicates success if the mean pose error is within the goal threshold $\beta = \frac{\pi}{12}$ and the \dkitty is sufficiently upright at the last step ($t = T$) of the episode:
    $$ \phi_{se}(\pi) = \mathbb{E}_{\tau\sim\mathcal{\pi}}[\mathbb{1}(\bar{e}_{T, pose}^{(\tau)} < \beta) *  \mathbb{1}(u_{T, kitty}^{(\tau)} > 0.9) ] $$
    
    \item \textbf{Orient}:
    This task involves \dkitty matching its current facing direction $\bm{\omega}_{t}$ with a goal facing direction $\bm{\omega}_{goal}$, thus minimizing the facing angle error $e_{t, facing}$ between $\bm{\omega}_{desired}$ and $\bm{\omega}_{t}$.
    The observation space is a 53-size 1D vector of the shared observation space entries, $\bm{\omega}_{t}$ and $\bm{\omega}_{goal}$ represented as unit vectors on the (X,Y) plane, and angle error $e_{t, facing}$.
    The reward function is defined as:
    \begin{align*}
        r_t &= r_{t, upright} - 4 {e}_{t,facing} - 4 || \bm{p}_{t, kitty} ||_2 + r_{bonus\_small} + r_{bonus\_big} \\
        r_{bonus\_small} &= 5 ({e}_{t,facing} < 15\degree \text{ or } u_{t, kitty} > \cos(15\degree)) \\
        r_{bonus\_big} &= 10 ({e}_{t,facing} < 5\degree \text{ and } u_{t, kitty} > \cos(15\degree))
    \end{align*}
    where the shared reward function constants are $\alpha_{upright} = 2$, $\alpha_{falling} = -500$, $\beta = \cos(25\degree)$.
    
    Three variants of this task are provided:
    \begin{enumerate}
        \item \textit{DKittyOrientFixed}: constant initial facing ($0\degree$) and goal facing ($180\degree$).
        \item \textit{DKittyOrientRandom}: random initial facing ($[-60\degree, 60\degree]$) and goal facing ($[120\degree, 240\degree]$)
        \item \textit{DKittyOrientRandomDynamics}: same as previous. The joint gains, damping, friction loss, geometry friction coefficients, and masses are randomized. In addition, a randomized height field is generated with heights up to $0.05\si{m}$
    \end{enumerate}
    
    The successor evaluator indicates success if the facing angle error is within the goal threshold and the \dkitty is sufficiently upright at the last step ($t = T$) of the episode:
    $$ \phi_{se}(\pi) = \mathbb{E}_{\tau\sim\mathcal{\pi}}[\mathbb{1}({e}_{T,facing}^{(\tau)} < 5\degree) * \mathbb{1}(u_{T, kitty}^{(\tau)} > \cos(15\degree)] $$
    
    \item \textbf{Walk}:
    This task has the \dkitty move its current Cartesian position $\bm{p}_{t,kitty}$ to a desired Cartesian position $\bm{p}_{goal}$, minimizing the distance $d_{t, goal} = || \bm{p}_{goal} - \bm{p}_{t, kitty} ||_2$. Additionally, the \dkitty is incentivized to face towards the goal. The heading alignment is calculated as $h_{t, goal} = \bm{R}_{\hat{y},t, kitty} \cdot \frac{\bm{p}_{goal} - \bm{p}_{t, kitty}}{d_{t, goal}} $. The observation space is a 52-size 1D vector of the shared observation space entries, $h_{t, goal}$ and $\bm{p}_{goal} - \bm{p}_{t, kitty}$.
    
    The reward function is defined as:
    \begin{align*}
        r_t &= r_{t, upright} - 4 d_{t, goal} + 2 h_{t, goal} + r_{bonus\_small} + r_{bonus\_big} \\
        r_{bonus\_small} &= 5 (d_{t, goal} < 0.5 \text{ or } h_{t, goal} > \cos(25\degree)) \\
        r_{bonus\_big} &= 10 (d_{t, goal} < 0.5 \text{ and } h_{t, goal} > \cos(25\degree))
    \end{align*}
    and the shared reward function constants are $\alpha_{upright} = 1$, $\alpha_{falling} = -500$, $\beta = \cos(25\degree)$.
    
    Three variants of this task are provided:
    \begin{enumerate}
        \item \textit{DKittyWalkFixed}: constant distance ($2\si{m}$) towards $0\degree$.
        \item \textit{DKittyWalkRandom}: random distance ($[1, 2]$) towards random angle ($[-60\degree, 60\degree]$)
        \item \textit{DKittyWalkRandomDynamics}: same as previous. The joint gains, damping, friction loss, geometry friction coefficients, and masses are randomized. In addition, a randomized height field is generated with heights up to $0.05\si{m}$
    \end{enumerate}
    
    The successor evaluator indicates success if the goal distance is within a threshold and the \dkitty is sufficient upright at the last step of the episode:
    $$ \phi_{se}(\pi) = \mathbb{E}_{\tau\sim\mathcal{\pi}}[ \mathbb{1}(d_{T, goal}^{(\tau)} < 0.5) * \mathbb{1}(u_{T, kitty}^{(\tau)} > \cos(25\degree))] $$

\end{enumerate}

\subsection{Safety metrics}

The following safety scores are shared between all tasks.

\renewcommand{\labelenumi}{\alph{enumi})}
\begin{enumerate}[wide, label=(\roman*), labelindent=-5pt, leftmargin=5mm]

    \item \textbf{Position violations}:
    This score indicates that the joint positions are near their operating bounds. For the $N$ joints of the robot, this is defined as:
    \begin{align*}
        s_{position} &= \sum_{i=1}^{N} \Big(\mathbb{1}(|\theta_i - \beta_{i,lower}| < \epsilon) + \mathbb{1}(|\theta_i - \beta_{i,upper}| < \epsilon)\Big)
    \end{align*}
    where $\beta_{i,lower}$ and $\beta_{i,upper}$ is the respective lower and upper joint position bound for the $i$th joint, and $\epsilon$ is the threshold within which the joint position is considered to be near the bound.
    
    \item \textbf{Velocity violations}:
    This score indicates that the joint velocities are exceeding a safety limit For the $N$ joints of the robot, this is defined as:
    $$ s_{velocity} = \sum_{i=1}^{N} \mathbb{1}(|\dot{\theta}_i| > \alpha_i) $$
    where $\alpha_i$ is the speed limit for the $i$th joint.
    
    \item \textbf{Current violations}:
    This score indicates that the joints are exerting forces that exceed a safety limit. For the $N$ joints of the robot, this is defined as:
    $$ s_{current} = \sum_{i=1}^{N} \mathbb{1}(|k_i| > \gamma_i) $$
    where $\gamma_i$ is the current limit for the $i$th joint.

\end{enumerate}

\section{Locomotion benchmark performance on \dkitty}
    \begin{figure}[H]
      \centering
        \includegraphics[width=1\linewidth]{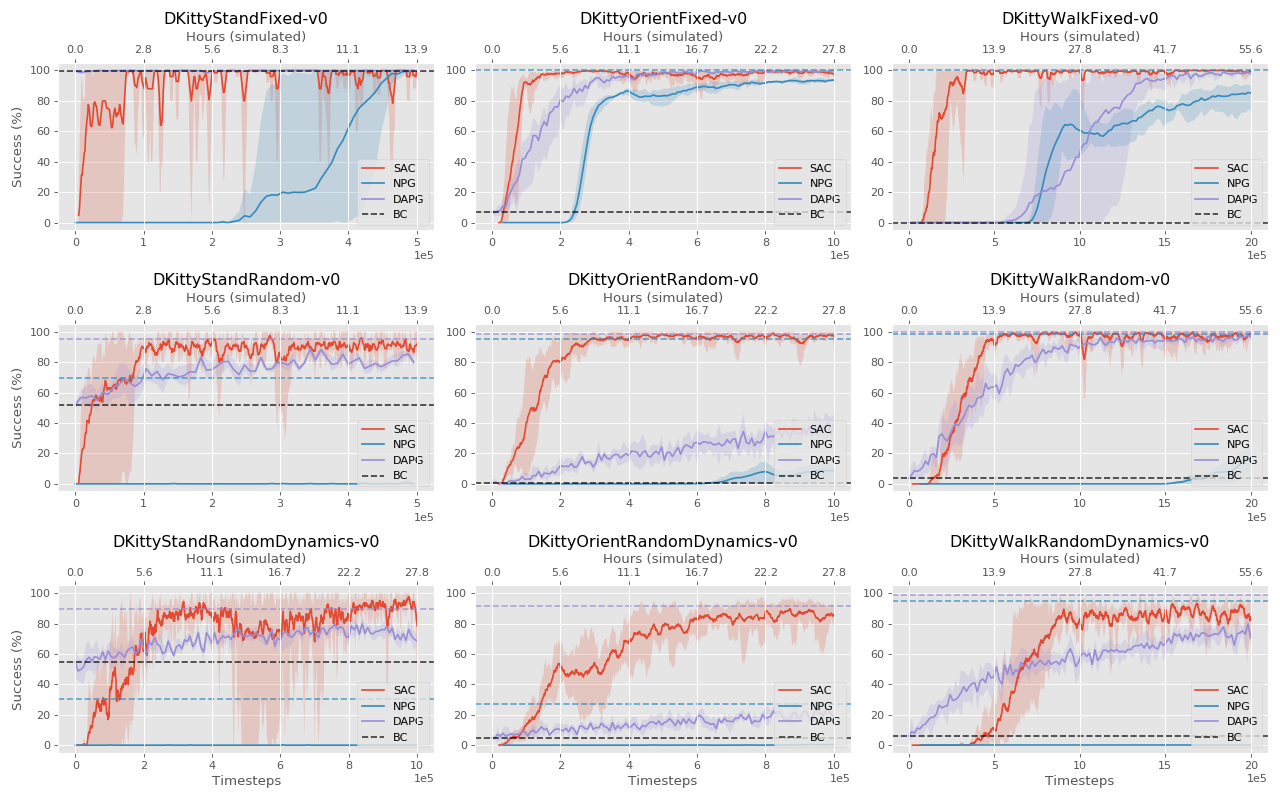}
        \caption{Success percentage (3 seeds) for all \dkitty tasks trained on a simulated \dkitty robot using  Soft Actor Critic (SAC), Natural Policy Gradient (NPG), Demo-Augmented Policy Gradient (DAPG), and Behavior Cloning (BC) over 20 trajectories. Each timestep corresponds to $0.1$ simulated seconds.}
      \label{fig:dkitty-baselines-all}
    \end{figure}
    
\section{\dsuite reproducibility}
    \begin{figure}[h!]
      \centering
        \includegraphics[width=1\linewidth]{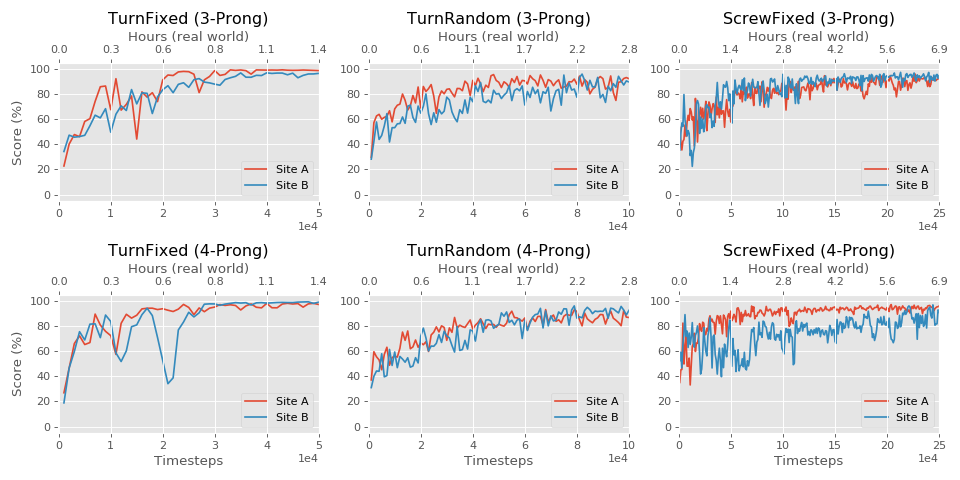}
        \caption{SAC training performance of \dclaw tasks on two real \dclaw robots each at different laboratory locations. Score denotes the closeness to the goal. Each timestep corresponds to $0.1$ simulated seconds. Each task is trained over two different task objects: a 3-prong valve and a 4-prong valve.}
      \label{fig:dkitty-baselines-all}
    \end{figure}

\end{document}